# ANALYSIS OF A STATISTICAL HYPOTHESIS BASED LEARNING MECHANISM FOR FASTER CRAWLING

Sudarshan Nandy[1] , Partha Pratim Sarkar[2] and Achintya Das[3]

[1] Department of Computer Science & Engineering, JIS College of Engineering, Kalyani, West Bengal, India.
`sudarshannandy@gmail.com`
[2] Partha Pratim Sarkar, DETS, University of Kalyani, Kalyani, West Bengal, India.
`parthabe91@yahoo.co.in`
Achintya Das[3], Kalyani Govt. Engineering College , Kalyani, West Bengal, India.
`achintya.das123@gmail.com`

## ABSTRACT

The growth of world-wide-web (WWW) spreads its wings from an intangible quantities of web-pages to a gigantic hub of web information which gradually increases the complexity of crawling process in a search engine. A search engine handles a lot of queries from various parts of this world, and the answers of it solely depend on the knowledge that it gathers by means of crawling. The information sharing becomes a most common habit of the society, and it is done by means of publishing structured, semi-structured and unstructured resources on the web. This social practice leads to an exponential growth of web-resource, and hence it became essential to crawl for continuous updating of web-knowledge and modification of several existing resources in any situation. In this paper one statistical hypothesis based learning mechanism is incorporated for learning the behaviour of crawling speed in different environment of network, and for intelligently control of the speed of crawler. The scaling technique is used to compare the performance proposed method with the standard crawler. The high speed performance is observed after scaling, and the retrieval of relevant web-resource in such a high speed is analysed.

## KEYWORDS

Focused Crawling, Crawling speed enhancement technique, statistical hypothesis,

## 1. INTRODUCTION

The reach content of web-resource makes the web more knowledgeable but at the same time it increases the time of crawling. The contents of a web is ranging from structured to unstructured data and it is represented by various web-pages which represent itself in the various format of semi-structured data to publish in the web. The Web itself is a big information or knowledge hub divided into several geographical locations. The job of a search engine is to search the query relevant web resources from the web. Search engine basically follows the several machine learning techniques to crawl the relevant web resource and another type of search engine only focuses for a particular domain of knowledge. Those are called domain specific search engine. The crawler collects the web services or the web resource from repository. The indexer job is to parse those web resources found by the crawler, and then relevant matches are submitted to the result interface [12]. The proposed method is incorporated in the standard focused crawling method for faster retrieval of web-resource. The empirical learning based on statistical hypothesis is also embedded in the system for adapting in dynamic environment of the web traffic while maintaining politeness in crawling speed. The network in the various part of the web can behave differently and in order to track that fluctuation in maintaining a stable mining speed one empirical learning method is applied.

## 2. REVIEW WORK

In recent years the development of web crawler is focused on the faster retrieval of web-pages deposition in less amount of time with a limited number of system resources. The system resources mainly used for this type of comparison is RAM and CPU usage during the running time of crawler. Some of the research focused on the above mentioned problem shows various ways to achieve its relevant answer. A few of those articles are targeted towards the resources-dependent crawling but they are not scalable [1,2,3,4,6,7,8,19]. The google search engine is one type of distributed search engine and uses multiple number of crawler to crawl the web[1,20]. The crawler designed at Kasetsart University in Thailand is reported a fast retrieval of web resources in '.th' domain[4]. The IRLbot is a large scale crawler and has a billion of web-resources in its knowledge hub. This crawler comes with new technique to deal with the reputation and spam problem. The Web resources may contain the unlimited number of host address and dump links for which the crawler may fall into an infinite loop and hence it clearly depicted the complex scenario where BFS scan is not only the possible solution to any crawler. So it is necessary to provide some decision making strategy based on real time observation to understand the web-resources content relevancy [9]. The IRLbot also reported in 2009 that the rate of crawling is 1789 pages/second in 319mb/s download speed [9]. One of the fastest crawler is reported in 2004 with 816 pages/sec speed, crawled more than 25 million pages [10]. Another crawler "AltaVista", in 2003 crawl more than 1.4 billion pages and contains 6.6 billion links [11]. The proposed method demands its ability to maintain politeness during crawling and also having its speciality by means of adaptive crawling through empirically learning the nature of that domain, and hence it is able to adjust the speed of crawling automatically in any network environment.

## 3. PROBLEM FORMULATION

In the proposed method the enhancement of crawling speed is achieved through a statistical hypothesis based learning on fluctuation of network bandwidth. On the basis of the knowledge of network bandwidth fluctuation the crawler robots are initiated on each iteration of a crawling level. In order to deploy parallel search to maintain a high speed in searching of a relevant web resources, each of the given node is considered as an independent hypertext graph. This hypertext graph is also an element of main graph G. So,

$$\left(G_1' \wedge G_2' \wedge G_3' \wedge \ldots \wedge G_N'\right) \in G$$

where $G_1'$, $G_2'$ etc. are different hypertext graphs and obviously belong to the different domains. In the initial stage of the experiment it is considered that all of those $G_N'$ contain the relevant as well as irrelevant web resources i.e. web resources under those graphs may be true or false from the focused search point of view. So it can be formulated in the way:

$$G_n' = \left( \sum_{(i=1 \in G_N')}^{(N)} V_i \wedge \neg \sum_{(i=1 \in G_N')}^{(N)} V_i \right)$$

$G_N'$ = the $N^{th}$ hypertext graph

$V_i$ = the i<sup>th</sup> web resource in the domain. It is also an element of the $G_N'$.
Now it is known that ,

$$G_N' \in G$$
$$So,$$
$$\left( \sum_{(i=1 \in G_N')}^{(N)} V_i \wedge \neg \sum_{(i=1 \in G_N')}^{(N)} V_i \right) \in G$$

and hence,

$$C_N \in G$$
$$where, C_N \subset G_N'$$

Some web resources are also unreachable from $G_N'$ and can be symbolically expressed as $U_{cn}$ and hence,

$$U_{cn} \subset G_N'$$

The time required to mine the all relevant web resources can be considered as T for crawling on $G_N'$. The speed at which it crawl the web resources is considered as $C_{si}$. In the dynamic environment of the web it is possible that the speed of the miner can fluctuate at any rate of any given running time in future or simply $\Box T$ . Now in a mine speed $C_{si}$, $C_n$ can be again true or false for a given set of focused topic and the false value of it can be represented by $\neg C_N$ Similarly in order to distinguish the high and low speed of mining, the relevant web resource the $C_{si}$ can be true or false and the false value can be represented by $\neg C_{si}$ . To understand the underlying logic and obviously the relation in between mine speed and its corresponding search of relevant or irrelevant web resources, those two parameters are divided into four states in a state transition machine with respect to the underlying representation of time.

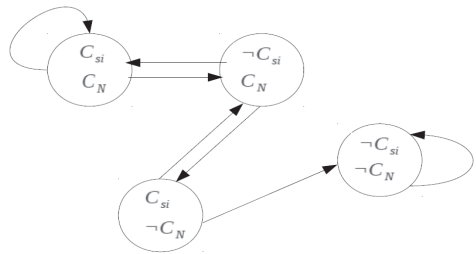

Figure 1. The state transition diagram.

The circle in the transition diagram represent the state, the arrows denotes the possible transition between the states. The states which may happen in case of the experiment are labeled with the literals. The literals some time represented with ' $\neg$ ' symbol to reflect that the meaning of the literals is "Not" in the same label for a given state. In the experiment it is indeed necessary to understand the truth value of the temporal formula

$$H = \Box C_{si} \vee \Box C_N$$

where H is calculated for each state of the figure 1.

i) H is true in $S_0$. States that are accessible from $S_0$ are $S_1$ and $S_0$ itself. Both $\Box C_{si}$ and $\Box C_N$ are true in $S_0$.

ii) H is true in $S_1$. States that are accessible from $S_1$ are $S_0$ and $S_2$. $\Box C_{si}$ is true in both the states. So the H is true.

iii) H is False in $S_2$. States that are accessible from $S_2$ are $S_1$ and $S_3$. Both of the literals are false in the $S_3$ and hence the H is also false.

iv) H is False in $S_3$. State that is accessible from $S_3$ is itself and hence it is also false.

The true value of H is needed to maintain at the running time of the crawler and the miner of the search engine will stop the work once the value of the H is false. In order to maintain the true value of H, it is also experimentally observable that how much less time it takes to visit maximum web resources of $G_N'$. It is also necessary to achieve high $C_{si}$ value in less amount of time and the same will be possible when the crawler will adaptively work in the dynamic environment of the web.

## 4. SYSTEM METHODOLOGY & ALGORITHMS

The proposed method consists of three parts, Master scheduler, responsible for controlling every process of the method but it is generally useful in case of handling all the relevant web resources and DB API, Indexer, extract all the necessary information from web resources and thereby maintain the request of the next focused resource used for crawling purpose and is recognized here as a IWM or Intelligent Web-Miner. All the important parts of the proposed system is depicted in figure 2. The master scheduler part assigns independent hypertext graph to every IWM. It generally put the requests of the all IWM in a schedule queue where the requests are handled using FIFO technique. On the other hand it also handles a process which inherits the statistical observation on crawling speed $C_{si}$, the error $\xi_{CS}$, harvest ratio of the method from the local DB which is maintained by master scheduler for storing the overall result of the crawling process. The master scheduler also helps to store the relevant URL in the 'URL_store' DB, and its corresponding document is stored in 'DOC_tab' DB.

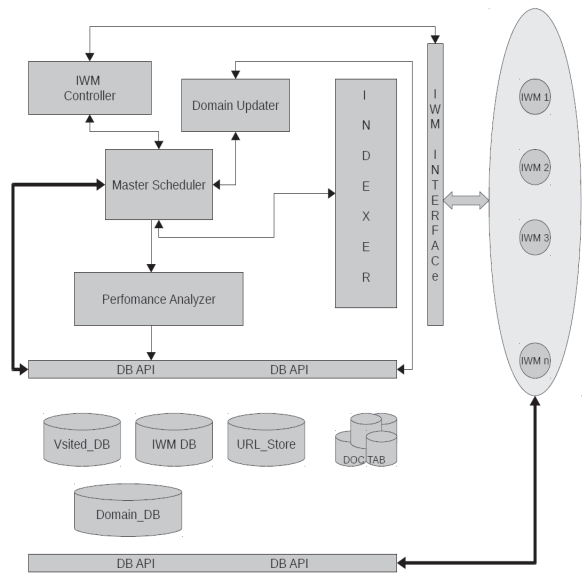

Figure 2. Architecture of proposed method.

The indexer is the most important part of the proposed system responsible for the extraction of all hypertext data and other necessary document from web resources. The web resources relevancy is mainly decided by this part of the system, and this decision is handled by a separate process, working under the Indexer, a daemon process. In case of retrieval of new relevant document, it sends all of it to the master scheduler, and then the master scheduler is responsible to transfer that document in proper place based on the decision coming from another process which is used to learn and recognize the matter of that web resources and the given query by using some machine learning technique.

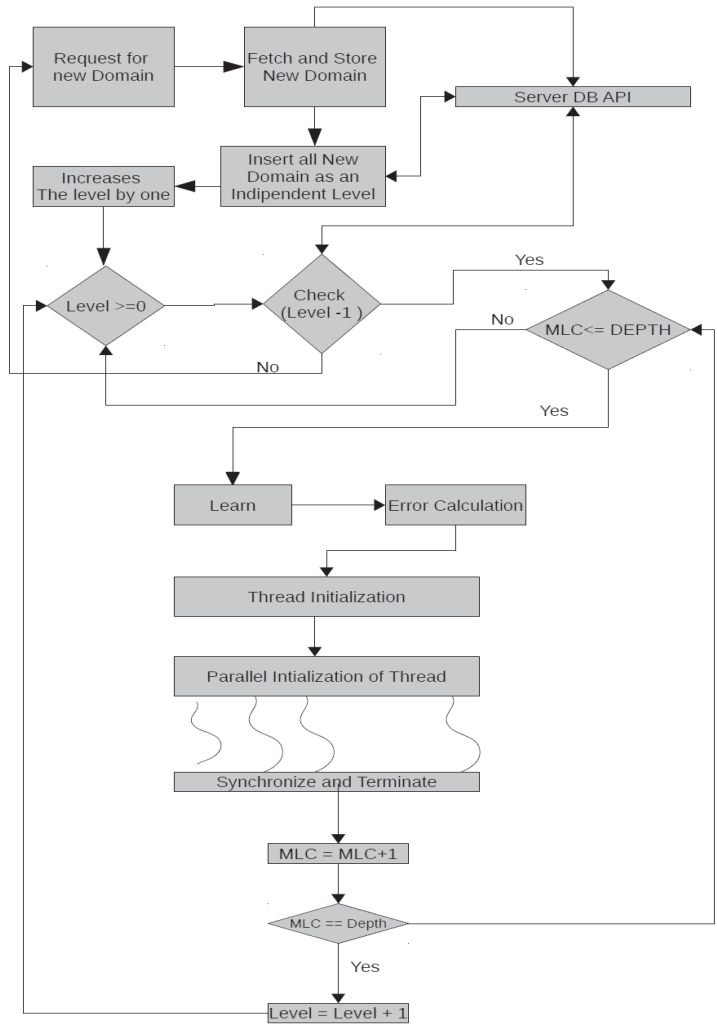

Figure 3. Block Diagram of IWM organizations .

Intelligent web-miner is one type of module used for crawling purpose. This module is actually responsible for crawling of web resources available from different domain. According to the proposed method this module can run on a different system with other similar type of module running on different system. So several modules can run parallel with different domains in searching of focused topic. Modules that run on different systems can be centrally controlled through the IWM controller block as in figure 2. Every module has its own learning technique to learn the environment on which several robots are working and those robots are actually used for crawling purpose. IWM of the proposed method has two goals; one is to crawl in a standard speed, and the other is to crawl the most relevant document. The architecture of the IWM is depicted in figure 3. as reference.

The algorithm of the IWM is as follows:

Algorithm of Intelligent Web-miner (IWM)

*Input: A set of URL_l contain 1<url_l<n ('n' is defined by the IWM controller) & relevant topic.
Output: A set of web focused resources .*

*Begin:*
    *1. Establish connection with server API.*
    *2. Set R_flag = 0*
    *3. While R_flag = 0:*
        *a. send a request for a set of URL*
        *b. If server sends a valid response then prepare local DB for receiving and storing data.*
    *c. On receiving data, set the R_flag = 1, and go to step 4, otherwise go to step3.*
    *4. Store each url of URL_l with different level tag, and store them for level wise initiation of hybrid crawling.*
    *5. A local depth named as $L_d$ is used to maintain hierarchical initiation of parallel crawling, because here parent of each hypertext child node is treated as independent hypertext graph of $L_{igh}$.*
    *6. increases $L_{igh}$ by 1*
    *7. While ($L_{igh}$ - 1) = valid:*
        *a. Append one hyper link of a $L_{igh}$ to a list.*
        *b. While $L_d <= D_{ms}$:*
            *i. Learn_cs($C_{si}$, $P_T$)*
            *ii. $P_T = C_T$    (where, $C_T >= 1$)*
            *iii. Fetch the child hypertext nodes ($h_n$) according to the number of $C_T$ where*
    *$hn \in Lihg$ ,and $h_n$ belongs to same level and same $L_d$ .*
            *iv. For each w in $h_n$ :*
                *1. Call ParallelCrawling ($W_n$)*
            *v. Synchronize and terminate all the light weight process vi.*
    *Increases $L_d$ by 1*
    *vii. If $L_d > D_{ms}$ :*
        *a. increases $L_{igh}$ by 1*
    *8. R_flag = 0 and Go to step 2.*

The IWM algorithm is used for the crawling purpose and designed in such a way that it can be plugged in to the main server at any point of running time of the system. The IWM procedure uses a level named $L_{igh}$ to indicate the number of each independent hypertext graph treated as parent or root. The number of $L_{igh}$ is depended on the availability of new focused domain found at the time of crawling, but it first starts with some defined focused domains. A local depth recognized as $L_d$ is used to maintain the depth of each hypertext child nodes (robots) generated from a single $L_{igh}$ . Now every child node or robot is treated as an independent graph, and the robots or threads are generated to crawl a specific web-resources . The decision on number of robot or thread is generated from the statistical hypothesis based learning method. The whole process is depicted in Fig. 4. Now, the algorithm of statistical hypothesis based learning method takes the current observation of $C_{si}$, the i[th] crawling speed, and number of $P_T$ or present thread (robot) .

Algorithm LearnCrawlingSpeed($C_{si}$, $P_T$)

Input: Crawl Speed $C_{si}$, Present no of used thread $P_T$.
Output: List of $E'_{cs}$ value ($L_{ecs}$), modified number of thread $P_T$.
Begin:
    1. a. $L_{ecs}$ = empty list, $L_f$ = 0 and
       b. learning limit or LL = n
    2. If $L_{ecs}$ <> n And $C_l$ <> LL And $L_f$ = 0 :
       i. Count the frequency of '0'
       ii. If frequency >= defined limit:
          a. Calculate current $C_s$
          b. $L_f$ = 1
          c. declare $L_{ecs}$ as an empty list
       iv. Store the $\xi_{CS}$, $C_{si}$, $P_T$ in local DB and goto Step 3.
    3. i. If $L_f$ = 0 and $L_{ecs}$ <= n:
       a. Calculate $\xi_{CS}$
       b. calculate the value of $E'_{cs}$ = $f(E'_{cs})$
       c. Deiced $P_T$ and append the $E'_{cs}$ value in $L_{ecs}$.
     ii. Else if $L_f$ = 1 and $L_{ecs}$ <> n then:
       a. Check current value of crawling speed.
       b. If $C_s$ <= threshold value then:
          a. $L_f$ = 0
          b. Goto Step 3.
     iv. Return ($L_{ecs}$, $P_T$)

The learning part of the system (figure 4) has a 'Critic' block, and it is used to check the fluctuation rate, even if the learning flag is set to off mode. It is also responsible to reset the learning, if the fluctuation rate of the crawling speed is lower than that of the previously observed value. $L_{ecs}$ is a list containing the result of the $E'_{cs}$, and the learning flag is set on or off according to the number of appearance of the '0', '1' and '-1' in the list. The decision of the $P_T$ is done according to the following rules:

    1. If $E'_{cs}$ = 1 then $PTnew$ = $PTold$ - 1
    2. If $E'_{cs}$ = 0 then $P_{Tnew}$ = $P_{Told}$
    3. If $E'_{cs}$ = -1 then $P_{Tnew}$ = $P_{Told}$ +1

123

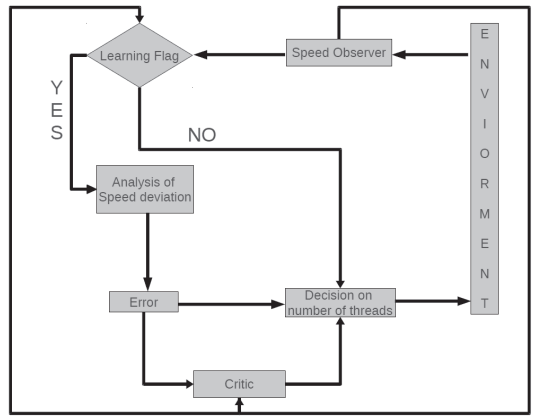

Finally the learning procedure returns the value of $P_T$ and $E_{cs}$ to the main IWM process. On receiving the $P_T$ value, the IWM procedure initiates that much number of crawling process for the parallel mining of the web resources from multiple number of domain. Then the IWM procedure synchronizes and terminates all the previously initiated crawling process to start with the new set of hypertext child node. This procedure of crawling ends, when $L_d$ reaches its limit ($D_{ms}$), which is defined by the main server process. R-flag is then set to '0', and IWM procedure requests and waits for response.

## 5. Evaluation

The performance of the proposed method is evaluated in terms of its crawling speed while it maintains politeness in crawling and relevance of crawled pages during the period of crawling. Intelligent Web-miner(IWM) is maintaining their own DNS cache and refresh its cache on certain time interval. The workloads of intelligent web-miner depends on the number of availability of new domain in crawler server data base and is distributed through master scheduler. If no new domain is available then master scheduler sends a wait signal to the other IWM's waiting queue. The crawler performance can be watched and configured at a visited page level based on the relevance score and crawling speed graph. The theoretical analysis and experiment on learning of crawl speed and page relevance at the same time are studied herewith in detail.

### 5.1 Experimental Set-up

The proposed distributed hybrid focused crawler is a developed python based application in this experiment where some C program is used as a plug-in. The method is developed on core 2 Duo processor 1.73 GHz Pentium-IV PC with 2 GB of RAM and a SCSI hard drive. The storage of database is centrally maintained on a Sun Server. Intelligent web-miner (IWM) is developed also and run from a core 2 Duo processor 1 GHz Pentium-IV PC with 3GB of RAM. All the machines are connected through LAN and LAN is connected to ISP using full-duplex 2 MB/Sec. bandwidth of a leased line network connection.

The crawler is initialized with two or three master topic and that is represented using two or three node as a independent hyper text graph to crawl. The responsibility of the crawler is to find out relevant pages on the mentioned topic while it maintains a good speed in crawling. The topics used as a test case is 'News'.

In order to represent the experiment some selected results from above topics are enumerated at here. The crawling speed on actual network bandwidth is mathematically scaled and the crawling speed is represented on high network bandwidth of 250 MB/s. The actual relevance of pages during crawling based on this proposed method is also studied.

### 5.2 Analysis of Experimental Result

According to the idea of web-resources retrieval in the proposed method, there are several intelligent web miners working in parallel with main server and Indexing server. Those web resources miners are named as Intelligent Web-miner(IWM). In case of understanding, the nature of crawling speed, it is necessary to estimate the amount of distribution from crawling speed log, where the crawling speed values are stored. In this proposed method some statistical measures are applied to analyse and presumed the nature of the speed in different domains with varying network speed. The nature of mining in different domain with a varying network speed, it is necessary to observe and analyse the dispersion or the degree of scatterness in the speed.

The scatterness in the crawling speed is observed to gain knowledge of fluctuation in network bandwidth. So, if the total deviation of $C_s$ watched at $\Delta t$ time and $C_s$ of $(\Delta t-1)$ time is large, then it can be considered as a large deviation of crawling speed. Now in case of the experiment, it must show the graph line going up, if the deviation is large and it behaves in opposite manner, when the deviation of crawling speed is low. The crawling speed can be calculated as follows:

Crawling Speed = (Number of visited node /Actual crawling time)

So,

$$C_s = \frac{\sum_{i=1}^{M} V_i}{C_t} \quad (1)$$

The crawling speed $C_s$ is the $i^{th}$ visited node of $V_i$, and it is actually divided by the actual crawling time.

Actual crawling time = (Present crawling time -starting time of IWM)

So, now the mean of the crawling speed can be calculated as follows:

$$\wp_{cs} = \frac{\sum (crawling\ speed)}{(Number\ of\ observation)}$$

So, $$\wp_{cs} = \frac{\sum_{i=1}^{M} C_{si}}{M} \quad (2)$$

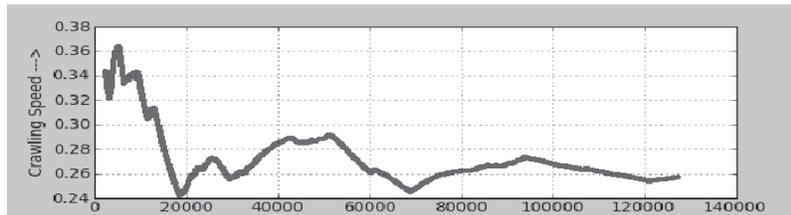

Figure 5. Fluctuation in Crawling Speed. ( x axis's is representing the actual crawling speed)

In the figure 5 the possible fluctuation rate of crawling speed can be observed. The mean square of the deviation of crawling speed is needed to know the behaviour of mining of web resources from a specific domain.
The mean square crawling speed can be calculated as follows:

$$Mean\ squre\ of\ Deviation\ C_{si}(\xi^2_{cs}) = \frac{\sum (no.of\ i^{th}\ crawling\ speed - mean\ of\ crawling\ speed)^2}{No\ of\ observation\ on\ crawling\ speed} \quad (3)$$

$$so, \ \xi_{cs}^2 = \frac{\sum_{i=1}^{M}(c_{si}-\wp_{cs})^2}{M} \quad (4)$$

$$or, \ M.\xi_{cs}^2 = \sum_{i=1}^{M}(c_{si}-\wp_{cs})^2$$

$$or, \ M.\xi_{cs}^2 = \sum_{i=1}^{M}\left(c_{si}^2 - 2c_{si}.\wp_{cs} + \wp_{cs}^2\right) \quad (5)$$

*Now, from (2) and (5) equation,*

$$M.\xi_{cs}^2 = \sum_{i=1}^{M} c_{si}^2 - 2\left(\frac{\sum_{i=1}^{M} c_{si}}{M}\right).\sum_{i=1}^{M} c_{si} + M.\left(\frac{\sum_{i=1}^{M} c_{si}}{M}\right)^2$$

$$or, \ M.\xi_{cs}^2 = \sum_{i=1}^{M} c_{si}^2 - \frac{\left(\sum_{i=1}^{M} c_{si}\right)^2}{M}$$

$$So, \ \xi_{cs}^2 = \frac{M.\left(\sum_{i=1}^{M} c_{si}^2\right) - \left(\sum_{i=1}^{M} c_{si}\right)^2}{M^2} \quad (6)$$

As now that standard deviation of any value is much more meaning full than that of mean square deviation, the standard deviation of crawling speed is calculated as follows:

$$\xi_{cs} = \sqrt{\frac{M.\left(\sum_{i=1}^{M} c_{si}^2\right) - \left(\sum_{i=1}^{M} c_{si}\right)^2}{M^2}} \quad (7)$$

According to the proposed method, it is always expected that at any crawling time, lower value of $\xi_{cs}$ is observed. It is also proved from the experiment that the lower value of $C_{si}$ always generates high value of $\xi_{cs}$.

In the proposed method, it is possible to plug many number of IWM with the main servers, because it is totally a distributed system. Each of these IWM work on different domains of web, and it is possible to plug it to the main servers from any area of the web. So those IWM will not work from same roof and hence it must face the fluctuation of network bandwidth caused by the network traffic of that area. One empirical learning based solution is used in the proposed method to apply the knowledge of network traffic fluctuation in every robot of an IWM. The effect of mining the relevant document in the available bandwidth at any time can be measured by calculating the $C_{si}$ or crawling speed, and the standard deviation of the same will be helpful to watch the small dispersion in the speed. So , in the experiment, it is always necessary to check the range of standard deviation using the following:

$$E_{cs} = \left[\left(\sum_{(i=0)}^{(M-1)} \xi_{csi}\right) - \xi_{csM}\right] \quad (8)$$

$E_{cs}$ is the difference between sum of all previously observed standard deviation of crawling speed and present standard deviation of the same. Clearly, the value of $E_{cs}$ belongs to the positive region, if the present mining speed of IWM is smaller than that of the previously observed value. If it belongs to the negative region, the value of mining speed or $C_{si}$ is in the standard level, and corresponding standard deviation of the mining speed is smaller than that of the previously observed $\xi_{csi}$ value. So to keep track on the values of $E_{cs}$, the following function is developed:

$$E_{cs}' = f(E_{cs}) \quad (9)$$

$$E_{cs}' = \begin{cases} -1 & \text{if } E_{cs} < 0.0 \\ 0 & \text{if } E_{cs} = 0.0 \\ 1 & \text{if } E_{cs} > 0.0 \end{cases}$$

Now the function $f(E_{cs})$ is used for separating all the negative and positive values and is finally put to those on a list for constant watching of the appearance of '0' value in the list. In case of the appearance of the zero value, the number of thread or the generation of robot is not increased or decreased; it will remain same for that particular iteration. If '1' appears on the list, it clearly depicts the dropping of $C_{si}$ for that particular iteration, when number of thread or robot decreases its value by one. The high crawling speed reflects when '-1' appears in the list and the number of thread or robot in this particular case increases by one for that iteration. The '0' value appearance in the list is good, because it reflects the stability of web mining in the existing network bandwidth. It is also observable that after certain high peak, mining speed is certainly dropped, and it becomes stable on some lower speed, also zero will appear in the list. So the question is how to tackle all such cases where crawling speed values are trapped on small range value. The proposed method consists of a crawling speed watcher, named here as a critic whose responsibility is to watch for any small or large fluctuation in $\xi_{cs}$ value. If any such case happens with this empirical based statistical learning method, critic is there to recover the mining speed or $C_{si}$ from that particular small range of crawling speed.

Figure 6 and 7 represent the proposed learning procedure. In figure 6 and 7 the first subplot represents the error rate indicated by the 'deep gray' line. The X axis represents the actual crawling time in second, and it is observed from the figure 6 and 7 that 'Error' or standard

International Journal of Artificial Intelligence & Applications (IJAIA), Vol.3, No.4, July 2012

deviation is plotted with high peak till 5000 second and after that it is being stabilized slowly. In the second subplot of the figure 6, crawling speed(y axis) is plotted with actual crawling time (x axis). The 'dark gray' line used to represent the crawling speed and it is observed that the crawling speed is going down to 5000 second but the 'critic' module in the proposed method correct it and it becomes stabilized from the 10000 second. The 'light gray' line in second subplot of the figure 6 and figure 7 represents the probable crawling speed on that time.

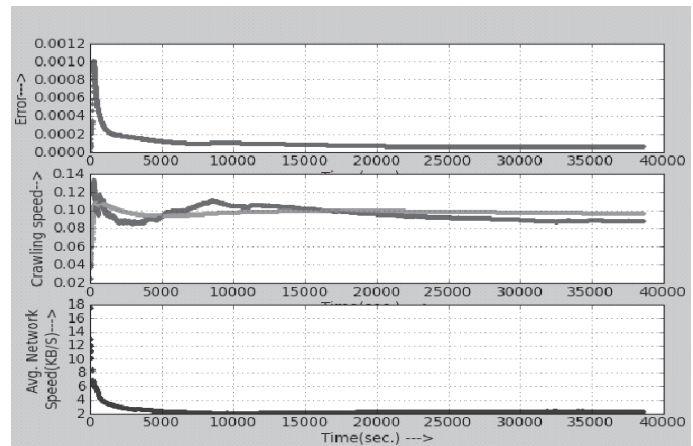

Figure 6. Empirical Learning based on statistical hypothesis.

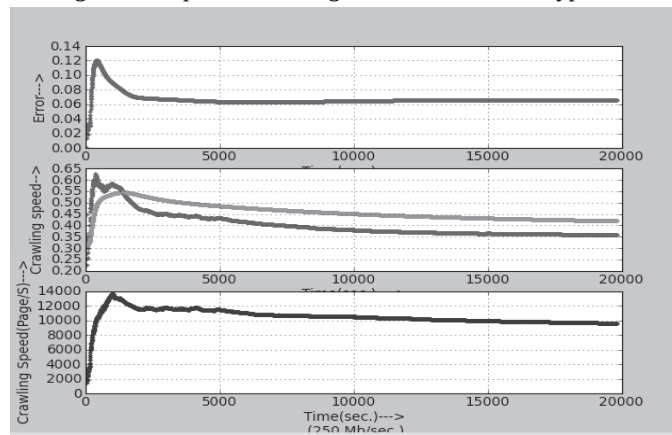

Figure 7. Scaling of crawling speed on 250mb/sec of network bandwidth.

In third sub-plotting of figure 7. the crawling speed is scaled to 250 mb/sec. of network bandwidth. On such plotting it is clearly observed from the figure 7. that after scaling the crawling speed varies from 10000 to 14000 pages/second. The actual crawling speed is depicted in second sub-plotting of figure 7. The actual network speed that is observed in figure 6 is 18 kb/sec at the initial stage of crawling and gradually decreases to 2 kb/sec. On scaling this network speed to 250 mb/sec it is found that crawling of pages can be mapped within the range of 10000 to 14000 pages/second (figure 7, third sub-plot).

In case of focused crawling, one of the most important point to understand is the relevance of the retrieved document on real time basis[15]. The document can be classified as a relevant one if the similarity can be found in between web resources and the focused topic [18, 14, and 16]. In this proposed method, the VSM based method [17] is used to classify in between various retrieved web-resources. This testing is essential to check whether the proposed method is able to retrieve focused web-resource or not. The crawling with the proposed method is started with a set link relevant to the 'News' topic, and it returns the links of the relevant web-pages . The links are listed below on the basis of top relevance:

| | |
|---|---|
| http://www.telegraph.co.uk/ | http://msn.com |
| http://wordpress.com/ | http://news.yahoo.com/ |
| http://www.bbcamerica.com/ | http://www.espn.co.uk/ |
| http://www.bbcworldnews.com/ | http://www.foxsports.com |
| http://www.nytimes.com/ | http://www.football.co.uk/ |
| http://www.motorcyclenews.com/ | http://espn.co.uk |
| http://reuters.com | http://politics.theatlantic.com/ |
| http://www.nypost.com/ | http://abcnews.go.com/ |

Those links are collected randomly from the huge collection of result which is based on relevant topic 'News'.

## 6. CONCLUSIONS

The proposed method is used to maintain the politeness at the time of crawling on a fluctuated network bandwidth. The crawling speed in the present is scaled from 10000 to 14000 pages/second in 250 mb/sec. , and it is based on one IWM. The proposed method is designed in such a way that any number of IWM can be used. The IWM can be plugged into the main server at any point of the crawling time. The proposed method maintains the focused crawling through a VSM based learning algorithm. The keyword similarity in between the document and query word is decided through cosine formula. The relevancy value of each document is observed in between zero and one and hence the efficacy of the present work gets established with an improvement in the crawling speed for distributed hybrid focused crawler.

## Authors

Sudarshan Nandy was born on 6thday of March 1983. He completed B.Tech in Computer Science and Engineering from Utkal University,and M.Tech in Computer Science and Engineering from West Bengal University of Technology in the years 2004 and 2007 respectively. He has been serving as Assistant Professor in JIS College of Engineering, since 2009. He is pursuing his Ph.D work under the guidance of Prof. Partha Pratim Sarkar and Prof. Achintya Das. His area of interest is Computational Intelligence, Web Intelligence, Meta-heuristics algorithm and Neural Network. 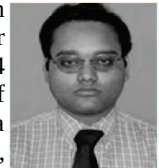

Dr. Partha Pratim Sarkar obtained his Ph.D in engineering from Jadavpur University in the year 2002. He has obtained his M.E from Jadavpur University in the year 1994. He earned his B.E degree in Electronics and Telecommunication Engineering from Bengal Engineering College (Presently known as Bengal Engineering and Science University, Shibpur) in the year 1991. He is presently working as Senior Scientific Officer (Professor Rank) at the Dept. of Engineering & Technological Studies, University of Kalyani. His area of research includes, Microstrip Antenna, Microstrip Filter, Frequency Selective Surfaces, and Artificial Neural Network. He has contributed to numerous research articles in various journals and conferences of repute. He is also a life Fellow of IETE. 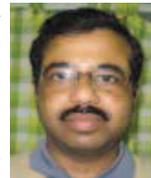

Dr. Achintya Das was born on 8th February 1957.He completed B.Tech., M.Tech and Ph.D (Tech) in the subject of Radio Physics Electronics from Calcutta University in the years of 1978, 1982 and 1996 respectively. He served as Executive Engineer in Philips from 1982 to 1996. He is Professor and Head of the department of Electronics and Communication Engineering of Kalyani Govt. Engineering College, Kalyani, West Bengal. He also worked for twelve years as Visiting Professor at Calcutta University. He has more than forty research publications so far. He is fellow members (life) of IE and IETE professional bodies. 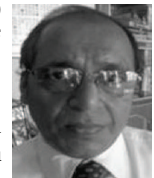